\documentclass{article}
\usepackage[margin=1in]{geometry}
\usepackage{amsmath,amssymb,amsfonts}
\usepackage{graphicx}
\usepackage{booktabs}
\usepackage[hidelinks]{hyperref}
\usepackage{natbib}
\usepackage{xcolor}
\usepackage{lmodern}
\usepackage[T1]{fontenc}
\usepackage{microtype}
\makeatletter
\newenvironment{recordtable}[1][12pt]
  {\par\addvspace{#1}\noindent
   \begin{minipage}{\linewidth}\def\@captype{table}}
  {\end{minipage}\par\addvspace{12pt}}
\makeatother
\title{Writability and the Pathway:\\
A Controlled Study of Knowledge Injection into Addressed Memory}

\author{%
  \begin{tabular}{@{}c@{\qquad}c@{}}
    Wei Gao\textsuperscript{1} & Cheng Tian\textsuperscript{1} \\
    {\small\texttt{gaowei@10rig.com}} &
    {\small\texttt{tiancheng2022@alumni.shanghaitech.edu.cn}} \\[0.8em]
    Duan Yang\textsuperscript{2} & Jianqing Yin\textsuperscript{1} \\
    {\small\texttt{yangduan@stu.xjtu.edu.cn}} &
    {\small\texttt{yinjianqing@10rig.com}}
  \end{tabular}\\[0.7em]
  {\small\textsuperscript{1}\,Shibei Qiyuan AI Technology}\\
  {\small\textsuperscript{2}\,Xi'an Jiaotong University}%
}

\date{October 2026}

\begin{document}
\maketitle

\begin{abstract}
Addressed conditional memory provides a local surface for post-hoc knowledge
injection, but a successful write need not transfer to a new question or
preserve untargeted behavior. This study combines a controlled GPT-2 comparison
with production-memory experiments on Qwen3.8-Flash-Next-FP8 and
DeepSeek-V4.1-Flash. Newly initialized modules attached to frozen GPT-2 are
tested in a $2\times2$ design: row-only versus row-plus-projection updates, each
with or without fixed module-residual anchoring. All arms reproduce 50 trained
completions in each of three sampling seeds. Natural-question target-string
hits average 5.3\% for both row-only arms, 28.7\% for unanchored projection
updates, and 22.0\% with anchoring, while strict answer scores remain zero.
Anchoring lowers measured drift; lexical hits do not establish entity-specific
retrieval. In the archived Qwen experiments, gradient writes make the
designated first answer token top-ranked for 132 independently reset targets,
comprising 93 new promotions and 39 already top-ranked targets. A separate
matched 24-target comparison gives 13 new promotions with Adam versus two with
SPSA at equal iteration counts but unequal computational procedures. An
80-row case includes saved continuations and a replay record, while
reformulations bound its scope. Six DeepSeek zero-order cases yield one new
target-token promotion and one regression. These results support bounded
writing feasibility and separate trained expression, question-form behavior,
and preservation; first-token rankings do not establish complete-answer success
or universal conditions for writable memory.
\end{abstract}

\section{Introduction}
\label{sec:intro}

Large language models acquire knowledge through pretraining on massive corpora,
while their pretrained weights remain fixed during ordinary inference. When
facts change or new information arrives, knowledge injection must address both
where to write that information and how the model will use it afterwards.
Conditional memory provides a substrate for local writes: table entries are
retrieved by deterministic addresses derived from the input token
sequence~\citep{cheng2026engram}.

This determinism suggests directly modifying the rows involved in new facts.
The parameter locations can be enumerated, but locality of the update does not
guarantee locality of its effects. Other inputs can reuse those rows, and the
retrieved vectors pass through shared projections and a context-dependent gate
before entering the backbone. Reproducing a trained completion therefore need
not establish that the same fact is accessible under a new question.

Prior work already demonstrates post-hoc memory writing. User as Engram studies
local row edits for personal memory; ENGRAFT uses multiple fact forms and
retention constraints; conditional memory adapters investigate retention-preserving domain
adaptation through conditional activation~\citep{li2026user,engraft2026,hou2026adapt}.
The remaining implementation choice examined here is whether writes should also
update the shared readout, and how to constrain the resulting drift. We study
these choices from a common mounted starting state, asking four connected
questions:
\begin{enumerate}
\item Which writing interface suffices to reproduce the trained completions:
addressed rows alone, or rows together with shared readout projections?
\item How do those writes support fact expression under an untrained natural-question form?
\item What collateral changes appear on untargeted text and unwritten entities?
\item How does residual anchoring affect these writing and preservation outcomes?
\end{enumerate}

We compare these choices in a $2\times2$ design: newly initialized memory modules
are attached to frozen GPT-2, and addressed-row or row-plus-projection updates
are crossed with the absence or presence of output-anchored writing (OAW).
OAW penalizes changes in module residuals relative to a fixed pre-write
reference. All arms share the same 50 canonical training sentences, complete
starting model, fact-sampling streams, and update count. Three seeds vary
optimization sampling on one initialization and fact set; anchor forward passes
make compute unequal despite equal update counts.

The study combines a controlled pathway comparison with an audit of fact
expression and collateral changes. Row-only writes already reproduce every
trained canonical completion at the tested budget. Shared projection updates
change question-form diagnostics and increase measured output drift; OAW lowers
drift within each scope while leaving canonical reproduction unchanged.
Complete per-seed measurements and saved continuations distinguish trained
completion, strict short-answer scores, lexical hits, and unwritten-entity
behavior. These results characterize the tested writing interfaces rather than
establishing universal prerequisites for writable memory.

The same questions matter for memory tables already trained inside production
models. We therefore retain the original Qwen and DeepSeek writing studies
alongside the controlled comparison (Section~\ref{sec:production}). Their
token-level outcomes, saved continuations, reformulations, and implementation
diagnostics provide complementary evidence, with their original protocols
reported separately. Earlier training-dose and controlled-bed observations
appear in Appendix~\ref{app:historical-beds}. Together, these studies examine
what can be written, how it is expressed, and which measurements delimit the
claim; they do not turn a successful completion into a universal writability
criterion.

\section{Related work}
\label{sec:related}

\paragraph{Post-hoc conditional-memory writing.}
User as Engram studies local parametric edits for per-user
memory~\citep{li2026user}. In its Joint OPT algorithm, a trainable tensor covers
the union of fact-triggered rows and gradients update that tensor. Our
row-plus-projection arm additionally trains the memory modules' shared key/value
weights and biases; joint optimization of multiple facts is therefore distinct
from joint optimization of rows and readout projections. ENGRAFT writes to an
existing n-gram memory table using a usage corpus with statements, questions,
and other forms of the desired information, alongside output-distribution and
row constraints~\citep{engraft2026}. Its single-trigger experiments also report
context-dependent transfer failures despite shared rows. In the controlled experiment, we train one canonical sentence per fact
and evaluate an untrained question form. These differences in updated objects,
training exposure, backbone, and scoring preclude a numerical ranking across
papers. The present study adds a shared-start factor comparison within a fixed
new-module setting to this established line of work.

\paragraph{Retention constraints.}
Conditional memory adapters pursue retention-preserving domain adaptation
through conditional activation~\citep{hou2026adapt}, jointly training memory,
projection, and gate parameters. The present experiment compares row-only and
row-plus-projection writes from a common start, crossed with fixed residual
anchoring. Learning without Forgetting uses
output-based distillation to preserve previous behavior~\citep{li2017lwf},
while EWC constrains parameter changes~\citep{kirkpatrick2017ewc}. Our penalty
instead matches each mounted module's output residual to its fixed pre-write
reference on anchor inputs. Because module residuals and final output
distributions are different objects, we measure both residual MSE and
final-distribution KL, along with language-model perplexity. A comparison with
matched output-KL regularization or replay would be needed to establish a
relative advantage for the residual penalty.

\paragraph{Model editing and evaluation.}
ROME and MEMIT rewrite feed-forward key--value
associations~\citep{meng2022rome,meng2023memit}; AlphaEdit uses null-space
constraints to preserve knowledge during editing~\citep{fang2024alphaedit}
(ICLR 2025), and MEND learns an edit operator~\citep{mitchell2022mend}.
GRACE studies lifelong editing with discrete key--value
adapters~\citep{hartvigsen2023grace}. These methods supply the broader editing
context but target different parameter objects from an addressed n-gram table.
User as Engram also discusses differences between lexical and model-judged
scores~\citep{li2026user}; our evaluation focuses on how such diagnostic
differences appear in the present controlled comparison. MQuAKE motivates
testing consequences that require composition of edited facts~\citep{zhong2023mquake};
the current experiment evaluates single-fact prompts.

\paragraph{Conditional-memory architectures.}
Engram-style designs~\citep{cheng2026engram,zhou2026tngram} and latent-space
Lngram~\citep{zheng2026lngram} motivate memory lookup as an architectural
component. We use an independent causal polynomial addressing implementation
in newly attached modules; its relationship to these production and latent-space
systems is architectural rather than a reproduction of their configurations.

\section{Background and problem formulation}
\label{sec:problem}

\paragraph{An address is not an answer.}
An addressed memory retrieves table rows using a deterministic key derived from
the recent input tokens. A shared readout then transforms those rows into a
contribution to the backbone. In the modules studied here, key and value
projections and a hidden-state-dependent gate form that readout
(Equation~\ref{eq:module}). Editing a row therefore changes a component of a
computation, rather than placing a natural-language fact in an independently
queriable database. A new formulation can change the addresses, the hidden state
used by the gate, or both. This is why reproducing a written completion and using
the same fact in a question need separate measurements.

\paragraph{Local parameters and shared pathways.}
Row updates are local to the selected parameter addresses. Their behavioral
effects can still extend to other inputs that reuse those addresses and to later
positions in a sequence. Updating the shared projections expands the direct
parameter surface further: even a row outside the write set can acquire a new
readout. We compare these two interfaces without assuming that the larger one
is needed for canonical fact expression or that either guarantees behavioral
isolation.

\paragraph{The writing problem.}
Given a fixed model with mounted memory modules and a set of synthetic fact
sentences, we optimize a specified memory parameter subset while freezing the
backbone. We examine the resulting canonical completions, an untrained question
form, unwritten-entity controls, and changes from the pre-write distribution.
These are separate operational aspects of writability, rather than a single
binary property or a claim of semantic internalization. The preservation
reference is the model with the new modules already mounted. OAW constrains the
modules' residual contributions on fixed anchor inputs; final-distribution drift
and perplexity are measured separately. The next section specifies the modules,
update surfaces, and objective used to study this problem.

\section{Writing protocol and controlled setting}
\label{sec:setting}

\paragraph{Backbone and mounted modules.}
The backbone is a publicly pretrained GPT-2 checkpoint
(revision \texttt{607a30d7})~\citep{radford2019gpt2}, used with its original
weights and configuration.
The memory-disabled conversion was checked against the native backbone on
fixed inputs. On the tested inputs, float32 differences satisfied the tolerances
reported in Appendix~\ref{app:conversion}. Two memory modules are inserted after the
zero-based blocks~3 and~8. Each module has 24 heads with $400{,}000$ rows per
head and $32$ dimensions per row. The first 12 heads use a causal 2-gram
addressing order and the last 12 a causal 3-gram order, so the mixed table
indexes a longer context than a single-token lookup. Each layer owns an
independent table.

\paragraph{Addressing.}
Let $z_t$ be the input token ID at position $t$. For zero-based head
$j\in\{0,\ldots,23\}$, the order $o_j$ is 2 for $j<12$ and 3 otherwise.
In layer $L\in\{3,8\}$, addressing is
\begin{align}
 q_0 &= L,\qquad
 q_{u+1} = (a_jq_u+z_{t-u})\bmod p_j
       \quad (u=0,\ldots,o_j-1),\\
 s_j(t) &= 400000j + (q_{o_j}\bmod400000).
\end{align}
For $t-u<0$, $z_{t-u}$ is the EOS ID 50256. The ordered modulus and multiplier
constants $(p_j,a_j)$ are shared by both layers and listed in
Appendix~\ref{app:hashconstants}. This independent causal polynomial lookup is
not a bitwise reproduction of a production XOR scheme. The addressed rows are
deterministically enumerated from the fixed training inputs.

\paragraph{Update scope.}
Two scopes are compared. \emph{Row-only} updates the embedding rows addressed by
the canonical training inputs and leaves the shared key/value projections
untouched. \emph{Row+projection} additionally trains the key and value weights
and biases of the memory modules. Because the projections are shared across
addresses while rows are address-local, the two scopes differ in more than
parameter count: a projection change can affect addresses that are never
addressed by the training inputs.

The row sets are computed from each complete canonical training input
(fact plus newline) after removing only its final target token, and the addressing
is deterministic, so the touched rows are enumerated rather than sampled. The two
mounted layers yield $7819$ and $7821$ rows respectively, i.e.\ $500{,}480$
row scalars are \emph{eligible to change}. The row+projection arm additionally
carries $2{,}362{,}368$ shared projection scalars. The tables themselves hold
$614{,}400{,}000$ scalars. Importantly, the implementation keeps the parameter,
gradient, and Adam moments as dense tensors over the full table and applies a
binary row mask to the embedding gradient before the update; it does not allocate
optimizer state only for the $500{,}480$ eligible row scalars.

\paragraph{Module and write objective.}
A mounted module turns the addressed rows into a hidden-state-sized
contribution as follows. Each of the 24 heads returns a 32-dimensional lookup
$e_i$; the module concatenates them into $e\in\mathbb{R}^{768}$ and computes
\begin{equation}
k = W_K e + b_K,\qquad v = W_V e + b_V,\qquad
g = \sigma\!\left(\frac{k^{\top} h}{\sqrt{768}}\right),\qquad
c = \alpha\, g \odot v,
\label{eq:module}
\end{equation}
with $\alpha=1$ and $h$ the block's hidden state; the block output becomes
$h \leftarrow h + c$. $W_K,W_V,b_K,b_V$ are the shared memory projections. The
two mounted layers hold separate parameters but are both initialized from the
same seed $12345$ with $\mathcal{N}(0, 0.02^{2})$ weights and zero biases.
The initialization is fixed across runs, including its dependence across layers.

\begin{figure}[!t]
\centering
\includegraphics[width=\linewidth]{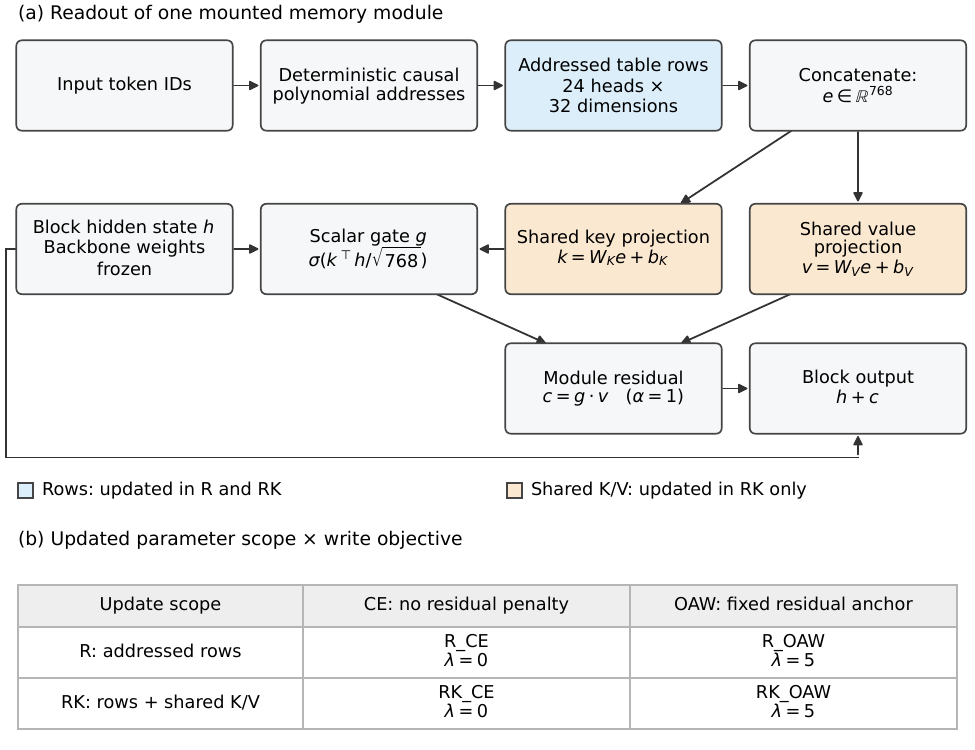}
\caption{Memory readout and the controlled write design.
(a) Causal token addresses select table rows; concatenated vectors pass through
shared key/value projections, a hidden-state-dependent scalar gate, and residual injection.
Separate modules are mounted after zero-based blocks 3 and 8.
(b) All arms update canonical-input addressed rows; RK also updates shared
key/value weights and biases. CE omits the residual penalty; OAW matches module
residuals to fixed pre-write references ($\lambda=5$).
Frozen backbone or projection weights do not imply fixed hidden states or gate activations.}
\label{fig:method-design}
\end{figure}

Figure~\ref{fig:method-design} summarizes the memory readout and the four update conditions.

At each writing step, the objective is
\begin{equation}
\mathcal{L} = 4 \cdot \widehat{\mathrm{CE}}_{\text{masked}}
\;+\; \lambda \cdot \frac{1}{2}\sum_{w=1}^{2}\sum_{\ell\in\{3,8\}}
\frac{\big\lVert c^{\theta}_{\ell}(x_w) -
c^{0}_{\ell}(x_w) \big\rVert^{2}_{F}}{T_w d},
\label{eq:objective}
\end{equation}
where $\widehat{\mathrm{CE}}_{\text{masked}}$ is the token-mean causal
cross-entropy over all real next-token positions of the sampled fact sentences.
The mask excludes only right-padding targets: subject, relation, answer, and
newline tokens are supervised wherever they occur as next-token targets. This
is a full-sentence language-model objective, not an answer-only loss. The
second term compares each module's actual forward contribution with its
fixed pre-write reference. The Frobenius norm sums over positions and features;
division by $T_w d$ gives the elementwise mean squared difference, with
$T_w=128$ and $d=768$. The term is averaged over the two sampled anchor windows
and summed over the mounted layers. Only the anchoring arms
evaluate the anchor term; the cross-entropy arms perform zero anchor forward
passes. The coefficient $\lambda$ is fixed at $0$ or $5$ by arm.

\paragraph{Experimental arms and optimization.}
We denote the row-only scope by R and the row-plus-projection scope by RK.
CE denotes the unanchored objective; OAW (output-anchored writing) denotes the
module-residual penalty in Equation~\ref{eq:objective}, not output-distribution
KL. Table~\ref{tab:arms} defines the four arms.

\begin{table}[htbp]
\centering\small
\begin{tabular}{llc}
\toprule
Arm & Updated parameters & $\lambda$ \\
\midrule
R\_CE & Addressed rows & 0 \\
R\_OAW & Addressed rows & 5 \\
RK\_CE & Addressed rows and shared K/V weights and biases & 0 \\
RK\_OAW & Addressed rows and shared K/V weights and biases & 5 \\
\bottomrule
\end{tabular}
\caption{The $2\times2$ design: update scope crossed with fixed residual anchoring.
The addressed-row set is identical across arms.}
\label{tab:arms}
\end{table}

Optimization uses Adam with row learning rate $0.005$, projection learning rate
$0.0005$, betas $(0.9, 0.999)$, epsilon $10^{-8}$, no weight decay and no gradient
clipping, in float32 with TF32 disabled and the model in evaluation mode
throughout. A run sees $50$ synthetic facts in batches of $8$ for $2500$ steps;
the write seeds are $42$, $777$, $1234$, and the shared base initialization seed
is $12345$. All four arms use the same update count and fact batches.
OAW adds two anchor windows per step; RK trains additional shared parameters.
The resulting computational costs are not matched.

\paragraph{Shared initial state and streams.}
Each run reconstructs the same complete pre-write model from the frozen
configuration. A content hash over backbone, buffers, and module parameters
verifies its identity before training. The source checkpoint, inputs, runner,
addressed-row sets, and full anchor reference are fixed for all arms. Fact and
anchor sampling streams are generated independently for each write seed and
shared across arms at that seed. At each step, a fresh random permutation
selects eight distinct facts from the 50-item pool and two distinct anchor
windows from the 63-window pool; items may recur across steps. Independent
generators use the write seed for facts and the write seed plus 100000 for
anchors. CE arms skip anchor forward passes without
altering their fact stream. Thus seed variation concerns optimization sampling
on one fixed initialization and dataset, rather than independently initialized
models or independently sampled fact sets.

\section{Evaluation}
\label{sec:evaluation}

\paragraph{Facts and prompt construction.}
We use a retained, fixed 100-item synthetic fact file. Items 0--49 are written;
items 50--99 are never trained and serve as controls. Each split has 50 distinct
subjects, with no subject shared across splits, and spans the seven relations in
Appendix~\ref{app:templates}. The splits contain 25 and 27 distinct target
strings, respectively, with 16 targets shared across them; subject separation
therefore does not imply answer separation. These records are synthetic
assertions rather than a benchmark of established world knowledge.
An answer-distribution diagnostic additionally assigns each example the most
frequent written target of its known relation, without using its subject or the
control answers to select that target. Its construction and scope are detailed
in Appendix~\ref{app:answer-distribution}.

For subject $s$, relation $r$, and answer $a$, training uses
\texttt{s + " " + r + " " + a + "\textbackslash n"}. The canonical prompt is
\texttt{s + " " + r}, with no trailing separator space. For example,
\texttt{Altherion was founded in the year 1380} is trained with a final newline,
and its canonical prompt is \texttt{Altherion was founded in the year}; the
expected continuation is \texttt{1380}. Canonical evaluation therefore tests a
prefix of the training sentence. The question form is generated from the fixed
relation template and is not included in training.

\paragraph{Response metrics.}
Normalized short-answer exact match measures whether the entire scored response
equals the expected answer under the declared normalization. The question
template does not enforce a short-answer-only response, so exact-match failures
may reflect answer formatting as well as factual errors. We therefore report
boundary-aware prefix and substring matches, target-sequence likelihoods, and
the underlying continuations. Substring matches alone do not establish a correct
factual assertion. The unmodified-backbone context control helps expose this
distinction under the same decoding and scoring protocol.

Four fifty-item sets are scored under one decoding and stopping rule:
(i) canonical completions of the written facts; (ii) a natural-question form of
\emph{the same} written facts, which is never trained; (iii) unwritten-entity
controls drawn from held-out synthetic entities; and (iv) the same natural
question form for those controls. The natural-question set is not fifty new
facts, and the control targets are not world knowledge. Raw continuations are
retained, and generation stops under a fixed rule; no result-dependent
truncation of the target length is applied.

\paragraph{Decoding and scoring.}
Generation is greedy with at most $32$ new tokens under a context limit of
$1024$; exceeding the context during generation raises an error rather than
sliding the window. The generated token sequence is stored with the end-of-text
token retained, and scoring excludes it. Scoring text is truncated before the
first newline character, so a token that itself contains a newline still
contributes the characters preceding it. Exact match is computed after the
declared normalization; prefix match is boundary-aware; substring match is
word-boundary aware. Target likelihood is teacher-forced over the jointly encoded
string \texttt{prompt + " " + target}, with token offsets used to locate the
continuation, so a bare numeric target is never encoded in isolation. The
question templates and normalization rules are listed in
Appendix~\ref{app:templates}; no few-shot examples or additional trained
question forms are introduced.

\paragraph{Reference model and fixed slices.}
Preservation measurements compare each written model with the common pre-write
model \emph{after} the new memory modules have been attached. They do not measure
the effect of mounting those modules relative to native GPT-2. We use three
disjoint slices of one recovered WikiText validation token stream
(Table~\ref{tab:slices}). A subsequent artifact audit exactly reconstructed the
stream from the raw validation text~\citep{merity2016pointer}, as described in
Appendix~\ref{app:artifact-identity}. The measurements concern these fixed slices
only. Anchor inputs occupy indices \texttt{[0,8192)}
and enter the OAW optimization objective. The development slice
\texttt{[8192,16384)} and perplexity slice \texttt{[20000,40000)} are not
used to optimize the writes. The anchor reference is captured before writing;
final preservation is measured on the separate development and perplexity slices.

\begin{table}[t]
\centering
\small
\begin{tabular}{llll}
\toprule
Slice & Window & Stride & Windows / positions \\
\midrule
Anchor (reference) & 129-token chunk & 128 & 63 windows, 128 inputs each \\
Development (drift) & 129-token chunk & 128 & 63 windows, 8064 positions \\
Perplexity (held out) & 513-token chunk & 512 & 39 windows, 19968 targets \\
\bottomrule
\end{tabular}
\caption{Fixed evaluation slices. The development KL is computed per position as
$D_{\mathrm{KL}}(\text{pre}\,\|\,\text{post})$ and averaged, in float64; the
perplexity is computed from accumulated negative log-likelihood over the
effective targets in the stated windows.}
\label{tab:slices}
\end{table}

A separate CPU address audit of this fixed prompt set found that, at the final
prompt position in each memory layer, all 24 head addresses belonged to the
written training-row union for every canonical prompt but for none of the
natural-question prompts; subject-disjoint controls are therefore not
address-disjoint controls (Appendix~\ref{app:address}).

Separate model evaluations measure each module's output residual MSE on the
development slice against its fixed pre-write reference. The native-backbone
context control uses the original, unaltered checkpoint. Unwritten-entity
controls do not represent facts that the pre-write model demonstrably knew,
so their exact match is not a forgetting measure.

For each control, the intrusion diagnostic considers distinct written target
strings from the same relation, excluding the control's own target. It applies
the same normalized, boundary-aware substring predicate as the Hit metric to
the scored continuation. A control with at least one such match is counted
once, with denominator 50, separately for canonical and question forms.
Own-target Hit and intrusion can both occur in one continuation; they are
different, non-complementary events, not semantic accuracy measures.

\section{Controlled GPT-2 results}
\label{sec:results}

The four questions lead from trained fact expression to question-form behavior,
preservation, and entity specificity. We first ask whether local row writes
suffice for canonical expression, then examine what changes with a shared
readout and with output anchoring. Table~\ref{tab:main} reports written-fact
outcomes; Tables~\ref{tab:pres} and~\ref{tab:ctrl} report drift and
unwritten-entity controls. All comparisons concern the same fixed starting model
and training exposure.

\begin{figure}[!t]
\centering
\includegraphics[width=\linewidth]{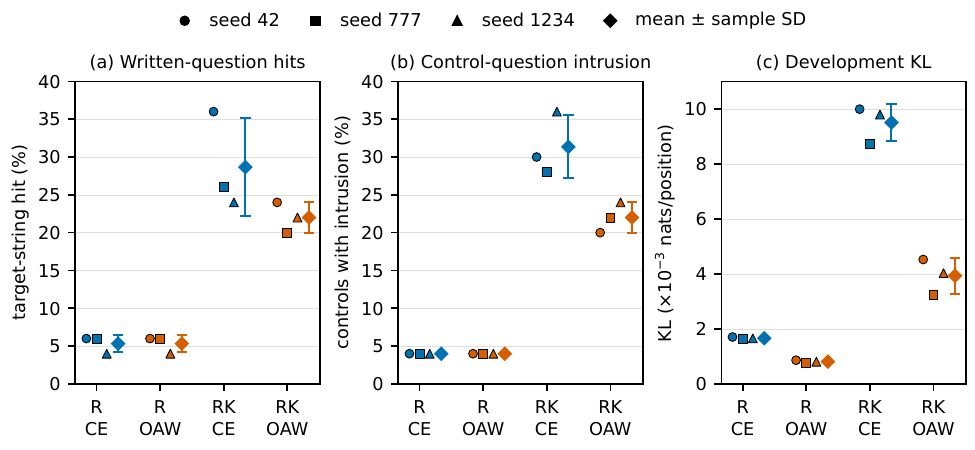}
\caption{Question-form diagnostics and output drift in the fixed GPT-2 setting.
R updates addressed rows; RK also updates shared key/value projections.
CE omits and OAW applies a fixed module-residual penalty.
(a) Boundary-aware target-string Hit on 50 written-fact questions.
(b) Intrusion on 50 unwritten-entity questions, matching another written target
of the same relation and excluding the control's own target; each control counts at most once.
(c) Pre$\|$post KL on 8064 development-slice positions.
Markers show three optimization-sampling seeds on one initialization and fact set;
horizontal offsets aid visibility. Diamonds and bars show mean$\pm$sample SD.
Canonical EM is 100\% in every run. Hit and intrusion are distinct lexical
events, not semantic accuracy.}
\label{fig:main-outcomes}
\end{figure}

Figure~\ref{fig:main-outcomes} summarizes the per-seed question-form diagnostics
and development-slice drift; the tables retain the exact outcomes and denominators.

\begin{table}[!htbp]
\centering\small
\setlength{\tabcolsep}{4pt}
\begin{tabular}{lcccc}
\toprule
Arm & Can.\ EM (\%) & Can.\ NLL & QA hit (\%) & QA NLL \\
\midrule
Common pre & $0.0$ & $5.05197$ & $0.0$ & $6.21891$ \\
R\_CE & $100.0\pm0.0$ & $(3.958\pm0.052)\times10^{-4}$ & $5.3\pm1.2$ & $5.16369\pm3.284\times10^{-3}$ \\
R\_OAW & $100.0\pm0.0$ & $(4.168\pm0.056)\times10^{-4}$ & $5.3\pm1.2$ & $5.19237\pm3.269\times10^{-3}$ \\
RK\_CE & $100.0\pm0.0$ & $(7.074\pm0.182)\times10^{-5}$ & $28.7\pm6.4$ & $4.78440\pm5.462\times10^{-3}$ \\
RK\_OAW & $100.0\pm0.0$ & $(7.048\pm0.172)\times10^{-5}$ & $22.0\pm2.0$ & $4.85989\pm9.761\times10^{-3}$ \\
\bottomrule
\end{tabular}
\caption{Canonical and natural-question outcomes. EM and boundary-aware target-string hit are percentages of 50 examples. NLL is the mean per-example target-token NLL, in nats. Entries are mean$\pm$sample standard deviation over three optimization-sampling seeds on one fixed model, initialization, and fact set. QA exact match and prefix hit are zero in all twelve runs. Target-string hits are content diagnostics, not semantic accuracy.}
\label{tab:main}
\end{table}
\paragraph{Fact expression through local row writes.}
All four arms achieve 100\% canonical exact match in every write seed, compared
with 0\% in the common pre-write state. Row-only updates therefore suffice for
these trained continuations at 2500 updates; changing shared key/value
projections is not necessary for this particular outcome. Mean canonical target
NLL is $3.96\times10^{-4}$ for R\_CE and $4.17\times10^{-4}$ for R\_OAW, versus
$7.07\times10^{-5}$ for RK\_CE and $7.05\times10^{-5}$ for RK\_OAW. This is a
measurement on the 50 synthetic training targets, not a capacity estimate or
evidence that the same result holds for pretrained memory modules.

\paragraph{From a written completion to a new question form.}
Natural-question exact match and prefix hit are zero in all twelve runs.
Boundary-aware target-string hits are $5.3\pm1.2$\% for both row-only arms,
$28.7\pm6.4$\% for RK\_CE, and $22.0\pm2.0$\% for RK\_OAW. Mean target NLL
falls from 6.219 before writing to 5.164, 5.192, 4.784, and 4.860, respectively.
Thus the strict short-answer score and the content diagnostics do not support
the same description of these continuations. Some responses contain the correct
target in a sentence. For example, for the Cindervale question in
Appendix~\ref{app:examples}, the projection arms name the written province
within a full statement, whereas the row-only arms name a different place;
all four fail strict short-answer EM. This example illustrates the scoring
distinction without estimating semantic accuracy over the full set. Neither zero exact
match nor a target-string hit alone establishes the presence or absence of
semantically correct transfer across the full set.

\begin{table}[!htbp]
\centering\small
\setlength{\tabcolsep}{4pt}
\begin{tabular}{lccc}
\toprule
Arm & Perplexity & PPL increase (\%) & Dev KL (nats/pos.) \\
\midrule
Common pre & $38.76495$ & --- & --- \\
R\_CE & $38.79852\pm0.00143$ & $0.0866\pm0.0037$ & $(1.663\pm0.041)\times10^{-3}$ \\
R\_OAW & $38.78303\pm0.00081$ & $0.0467\pm0.0021$ & $(8.156\pm0.453)\times10^{-4}$ \\
RK\_CE & $38.89841\pm0.00285$ & $0.3443\pm0.0074$ & $(9.504\pm0.678)\times10^{-3}$ \\
RK\_OAW & $38.84746\pm0.00376$ & $0.2129\pm0.0097$ & $(3.930\pm0.648)\times10^{-3}$ \\
\bottomrule
\end{tabular}
\caption{Preservation, as mean$\pm$sample standard deviation. PPL uses 39 windows and 19968 targets; each relative change uses the common pre-write value. Dev KL is pre$\|$post in float64 over 63 chunks and 8064 positions. Per-layer residual MSE appears in Appendix~\ref{app:runs}.}
\label{tab:pres}
\end{table}
\paragraph{Collateral drift and output-anchored writing.}
At the fixed penalty coefficient, mean development-slice KL changes from
0.001663 to 0.000816 nats per position in the row-only scope and from 0.009504
to 0.003930 in the row-plus-projection scope. Per-layer residual MSE also
decreases in both scopes. Relative to the common pre-write perplexity of 38.76495,
mean per-seed perplexity increases are 0.0866\%, 0.0467\%, 0.3443\%, and
0.2129\% for R\_CE, R\_OAW, RK\_CE, and RK\_OAW. The absolute increases are
small on this recovered slice. In every matched seed, anchoring lowers the
measured KL and perplexity within its scope, while natural-question target NLL
increases; in the projection scope the target-string hit rate also decreases.
At the tested coefficient, reduced output drift accompanies higher
question-form target NLL, and, in the projection scope, fewer target-string
hits. This is a descriptive comparison of the measured outcomes, not evidence
of a tradeoff in semantically correct retrieval across a wider population.

\begin{table}[!htbp]
\centering\small
\setlength{\tabcolsep}{4pt}
\begin{tabular}{lcccc}
\toprule
Arm & Can.\ EM (\%) & QA EM (\%) & Can.\ intr.\ (\%) & QA intr.\ (\%) \\
\midrule
Common pre & $0.0$ & $0.0$ & $0.0$ & $0.0$ \\
R\_CE & $29.3\pm1.2$ & $0.0\pm0.0$ & $68.7\pm3.1$ & $4.0\pm0.0$ \\
R\_OAW & $29.3\pm1.2$ & $0.0\pm0.0$ & $68.0\pm0.0$ & $4.0\pm0.0$ \\
RK\_CE & $18.7\pm2.3$ & $0.0\pm0.0$ & $81.3\pm2.3$ & $31.3\pm4.2$ \\
RK\_OAW & $18.7\pm1.2$ & $0.0\pm0.0$ & $81.3\pm1.2$ & $22.0\pm2.0$ \\
\bottomrule
\end{tabular}
\caption{Unwritten-entity controls, mean$\pm$sample standard deviation. Every percentage has denominator 50 controls. Intrusion counts each control at most once when it matches another written target of the same relation, excluding its own target. Control EM is not a forgetting score for facts the pre-write model demonstrably knew. Control NLL and paired changes appear in Appendix~\ref{app:runs}.}
\label{tab:ctrl}
\end{table}
\paragraph{Unwritten-entity responses and specificity.}
Canonical exact match on the unwritten synthetic controls rises from 0\% before
writing to $29.3\pm1.2$\% in each row-only arm and $18.7\pm2.3$\% and
$18.7\pm1.2$\% in RK\_CE and RK\_OAW. This is not a forgetting score for facts
the pre-write model demonstrably knew. Canonical intrusion rates average 68.7\%,
68.0\%, 81.3\%, and 81.3\%, respectively: each intrusion is a generated match
to at least one other written target of the same relation, excluding the
control's own target. Natural-question intrusion rates average 4.0\%, 4.0\%,
31.3\%, and 22.0\%. Target reuse and final-position address overlap limit
interpreting control exact match as entity-specific knowledge. In the projection
scope, written-question hits (28.7\% without anchoring and 22.0\% with it)
and unwritten-question intrusions (31.3\% and 22.0\%) change in the same
direction. Because these diagnostics match different target sets and events,
their rates cannot be subtracted to obtain a net accuracy or used to classify
all hits as leakage. They nevertheless show why higher question-form hits alone
do not establish entity-specific retrieval. The controls measure changes in
unwritten-entity behavior, not preservation of previously mastered facts.
Additional answer-distribution diagnostics in Appendix~\ref{app:answer-distribution}
quantify target concentration and distinguish written targets that are absent
from the control answer pool. These remain lexical measurements.

\section{Writing into production memory tables}
\label{sec:production}

The controlled experiment isolates update scope and residual anchoring in new
modules. Existing pretrained memory poses a complementary question: can local
table changes alter a production checkpoint's response without retraining its
backbone? The original experimental campaign included
\texttt{Qwen3.8-Flash-Next-FP8} and \texttt{DeepSeek-V4.1-Flash}, whose public
releases contain n-gram memory components~\citep{qwen2026flashnext,deepseek2026flash}.
This section reports their retained experimental records after checking the
saved measurements against the available writing code. It is an analysis of
those runs, not a fresh production-model replay. The protocols differ from the
GPT-2 experiment and do not test production-scale OAW.

\subsection{Local writes and the measured target}
\label{sec:production-protocol}

The Qwen implementation addresses rows through the checkpoint's n-gram lookup,
uses a custom multi-device loader, and writes into FP8 storage. A prompt's row
union is formed and rows addressed at position zero are excluded. For the
gradient-based procedure, a trainable additive vector is inserted at the
selected embedding lookups, using a pass-through backward implementation for
quantized operations. Adam optimizes this vector; it is then written back with
the stored scale and FP8 clamping. The table is restored after each target.
The comparison SPSA procedure instead evaluates quantized perturbations during
optimization. These are distinct computational procedures even at equal
iteration counts; the pass-through implementation does not establish exact
derivatives of a discrete quantization operation.

For each supplied answer string $a$, the scripts select its first tokenizer
token $a_0$ and measure its next-token rank after the prompt. We call rank zero
\emph{target-token top-1}. A \emph{new promotion} requires a nonzero baseline
rank and rank zero after writing. A target that was already top-ranked is
reported separately. These scores test a designated token, which may be an
answer prefix or whitespace, rather than a complete factual answer. Five
post-write argmax tokens in the 24-target gradient batch and one in the
132-target batch decode to whitespace. Consequently, neither a top-1 count nor
the supplied target labels establish full-answer accuracy.

\begin{table}[!ht]
\centering\small
\setlength{\tabcolsep}{4pt}
\begin{tabular}{llrrrrr}
\toprule
Checkpoint / set & Writer & Iter. & Pre top-1 & Post top-1 & New & Lost \\
\midrule
Qwen / matched 24 & SPSA & 150 & 11/24 & 13/24 & 2/13 & 0/11 \\
Qwen / matched 24 & Adam & 150 & 11/24 & 24/24 & 13/13 & 0/11 \\
Qwen / pool 132 & Adam & 60 & 39/132 & 132/132 & 93/93 & 0/39 \\
DeepSeek / 6 & SPSA & 150 & 3/6 & 3/6 & 1/3 & 1/3 \\
\bottomrule
\end{tabular}
\caption{Archived production target-token outcomes. New uses initially
non-top-1 targets as its denominator; Lost uses initially top-1 targets.
Targets are independent write--measure--restore trials, not a pool of
simultaneously retained edits. Iteration counts do not match compute or query
budgets.}
\label{tab:production-summary}
\end{table}

\subsection{Qwen: write feasibility and optimizer dependence}

On the same 24 prompts and supplied answers, the saved SPSA and Adam records
match on the selected row count and the entire recorded baseline measurement.
At 150 iterations, Adam produces 13 additional target-token top-1 outcomes,
whereas SPSA produces two (Table~\ref{tab:production-summary}). The remaining
11 targets were already top-ranked. This comparison shows that a failure to
promote a target under the tested SPSA procedure does not establish that the
table cannot support that token-level write. It does not rank the optimizers
under equal computation, and the different optimization-time representations
prevent attributing the difference solely to access to a derivative.

The larger batch contains 132 separately restored target-specific trials.
Adam uses 60 iterations per target, promoting all 93 initially non-top-1 target tokens and retaining the
39 already top-ranked tokens within their respective trials. The deepest
promoted token starts at rank 2,153. Mean log-probability gain is 4.10 nats over
all 132 targets and 5.46 over the 93 new promotions; the saved clipping counts
sum to zero. Because rows are restored after each target, these results do not
measure joint storage capacity, sequential retention, or transfer to new
questions. They establish the narrower observation that these designated
next-token outputs can be changed through local memory writes in the tested
production implementation.

\subsection{An 80-row write, reformulations, and controls}

For \texttt{The chemical symbol for gold is}, a saved write changes the target
\texttt{Au} from rank 1 to rank 0. The script excludes the 16 rows also addressed by four predefined control
prompts from a 96-row target union and writes the remaining 80. The stored short continuation changes
from \texttt{derived from its Latin} to \texttt{Au, derived from its Latin}.
A separate saved verification output agrees on the target log-probability
change, from $-1.51251$ to $-0.67533$, and reports zero maximum change in
next-token log-probabilities for four declared controls. These controls informed the write-set construction: their zero change checks
the designed locality, not preservation on unseen controls or a fresh replay.

The six saved reformulation and entity-control probes include short generated
continuations (Appendix~\ref{app:production-details}). The possessive form
already produces \texttt{Au} before writing, and the preposition variant's
continuation already contains it. The question-form snippet remains an
incomplete response prefix. Thus the records support neither ``all paraphrases
fail'' nor a general question-answering success rate. All six probes overlap
the write set, so they also do not test a zero-overlap reformulation. With the designed shared-row exclusion, the 80-row verification records zero
changes on the four controls; a different saved write changes three control
distributions. These are separate runs, not a matched preservation comparison.

\subsection{DeepSeek: mixed outcomes and implementation limits}

The six DeepSeek SPSA trials include one new target-token promotion, two
initially top-ranked targets that remain top-ranked, one regression from rank
0 to 2, and two targets that remain non-top-1. The promoted supplied target is
\texttt{Rio} after a Brazil-capital prompt; it is an imposed target, not evidence
that the resulting answer is factually correct. These positive and negative
outcomes show a limited ability to perturb the designated response, rather
than a general cross-architecture success rate.

Separate gradient diagnostics record zero gradients on the intervention tensor
while a finite perturbation changes the measured loss by 0.75532. However,
the gradient-batch baselines differ from the SPSA baselines, and the available
instrumented script appends the target token while its training loss uses the
final output position. Prediction-position alignment and the patched forward
therefore require further validation. The position log also flattens the
checkpoint's four residual branches; its entries cannot be read as independent
token positions without that mapping. These diagnostics document difficulties
in this implementation, not a universal obstruction caused by quantized
attention kernels. They do not establish that first-order writing is impossible
on DeepSeek or that SPSA is a necessary method.

\section{Discussion and limitations}
\label{sec:limits}

\paragraph{Writability and the readout pathway.}
The controlled comparison separates trained fact expression from the other
consequences of a write. Local rows suffice for canonical reproduction at the
tested budget; shared projection training changes question-form likelihood,
lexical diagnostics, and distributional drift. OAW reduces drift without
changing canonical exact match, while its question-form effects depend on update
scope and metric. Writability in this setting is therefore described by several
measured outcomes rather than a single completion score.

\paragraph{Fact expression and specificity.}
Canonical success conceals differences under new questions and on unwritten
entities. Exact match is sensitive to response format, while a target-string
hit need not assert the right fact about the right subject. Intrusion and
own-target Hit can co-occur. The saved continuations and complementary metrics
support this broader audit; quantifying correct, entity-specific retrieval
would additionally require semantic assessment of the responses.

\paragraph{From controlled beds to production memory.}
The production results complement the controlled comparison without supplying
the same intervention. They retain trained tables, different addressing and
readout implementations, and token-level objectives. The matched Qwen target
sets help distinguish feasibility from the limitations of one writing
procedure; they do not establish compute-normalized superiority. The
DeepSeek regression and reformulation probes preserve adverse outcomes rather
than treating every local response change as knowledge acquisition.

\paragraph{Experimental limits.}
The RK intervention changes both parameter count and cross-address sharing,
and OAW adds compute, so these comparisons do not isolate a single mechanism.
Final-position address analysis does not exclude retrieval through earlier
positions or hidden-state-dependent gating. The controlled experiment fixes one backbone,
one module initialization, one synthetic fact set, and one penalty coefficient;
three seeds vary only optimization sampling. Preservation is measured against
the mounted pre-write model on fixed slices of one WikiText validation stream.

Independent initializations and fact sets, additional preservation data, and
matched output-KL or replay controls would test the reach of these findings.
The production studies use different protocols and mostly first-token targets;
they neither test OAW nor provide multi-hop composition measurements. Pathway width, backbone scale, and prior training strength are not
isolated here; the results support neither a general three-condition criterion
nor method superiority.

\section{Reproducibility}
\label{sec:reproducibility}

The frozen protocol identifies the checkpoint, tokenizer, facts, slices, runner,
and sampling streams by content hash. Every run verifies the common pre-write
state and full anchor reference before optimization. Complete per-seed results
appear in Appendix~\ref{app:runs}; prompt construction and scoring are specified
in Section~\ref{sec:evaluation} and Appendix~\ref{app:templates}.
Memory-disabled conversion checks and their tolerances are reported in
Appendix~\ref{app:conversion}.

A checksum-verified private review bundle contains the facts, frozen code and
configuration, tokenizer, recovered token array, and all twelve saved result
records. These permit text-based rescoring and arithmetic checks of stored token
NLL values. Recomputing likelihoods, output KL, or perplexity requires model
states and forward passes; the presence of saved post-write state files alone
does not establish their reproduction.

The retained 247,290-token stream was exactly reconstructed from fixed raw
WikiText validation text. Appendix~\ref{app:artifact-identity} gives the public
revision and encoding procedure. This subsequent audit establishes artifact
identity; the historical processing program, formal dependency versions, and
synthetic-fact creation process remain unresolved. Materials are currently
private, and content hashes do not make them externally accessible.

For the production studies, the original result JSONs, current snapshots of the
writing/loader/FP8 code, saved case continuations, and available write-vector
metadata are retained separately. Counts in Section~\ref{sec:production} are
recomputed from those records, not obtained by new production inference.
Appendix~\ref{app:production-details} distinguishes saved replay outputs from
fresh reproduction and describes unresolved historical revision and prediction
alignment issues. Production and controlled protocols are not merged.

\begin{samepage}
\section{Conclusion}
\label{sec:conclusion}

Knowledge injection into addressed memory involves both a local intervention
and the model's subsequent use of that intervention. In the controlled GPT-2
experiment, addressed-row updates alone learn all trained canonical
completions. Shared projection updates change question-form diagnostics and
increase measured output drift; residual anchoring lowers drift while leaving
canonical reproduction unchanged.

The production studies preserve a complementary observation from the original
campaign: selected outputs of existing Qwen and DeepSeek memory can be changed
through local table writes. The Qwen records support designated first-token
promotions, a matched procedural comparison, and a saved 80-row response
change. The DeepSeek cases contain both a new promotion and a regression,
while its gradient diagnostics remain implementation-specific. These results
establish bounded feasibility, not simultaneous storage of all tested facts,
complete-answer accuracy, or production-scale OAW superiority.

Across the experiments, writability is best evaluated by reporting the update
surface, prediction target, transfer probes, and preservation measurements
together. Trained expression, expression under a new question, and behavior
on untargeted inputs remain separate outcomes. The retained evidence supports
these distinctions without a universal three-condition or scale-independence
claim.\par
\end{samepage}

\section*{Use of AI-assisted tools}
OpenAI Codex and DeepSeek Harness (using \texttt{deepseek-v4.1-flash}) were used
during this research and manuscript preparation.

\appendix

\section{All runs and per-seed metrics}
\label{app:runs}
All twelve final runs at seeds 42, 777, and 1234 are reported without selection.
\emph{Mean/std} summarizes those three runs using the sample standard deviation.
Percentages use 50 examples per set; target NLL is the mean per-example,
per-target-token NLL in nats. Seeds vary optimization sampling on one fixed
model, initialization, and fact set; no significance test is performed.

\begin{recordtable}[0pt]
\noindent\textbf{Canonical completions.}\par
Table~\ref{tab:seed-canonical} shows that every run reproduces all trained
completions. Target NLL separates scopes even after the discrete scores saturate;
these are measurements on the fixed training targets.
\par\smallskip
\centering\small
\setlength{\tabcolsep}{4pt}
\begin{tabular}{lccccc}
\toprule
Arm & Seed & EM (\%) & Prefix (\%) & Hit (\%) & Target NLL \\
\midrule
Common pre & --- & $0.0$ & $0.0$ & $0.0$ & $5.05197$ \\
R\_CE & 42 & $100.0$ & $100.0$ & $100.0$ & $3.958\times10^{-4}$ \\
R\_CE & 777 & $100.0$ & $100.0$ & $100.0$ & $4.009\times10^{-4}$ \\
R\_CE & 1234 & $100.0$ & $100.0$ & $100.0$ & $3.905\times10^{-4}$ \\
R\_CE & mean/std & $100.0\pm0.0$ & $100.0\pm0.0$ & $100.0\pm0.0$ & $(3.958\pm0.052)\times10^{-4}$ \\
R\_OAW & 42 & $100.0$ & $100.0$ & $100.0$ & $4.178\times10^{-4}$ \\
R\_OAW & 777 & $100.0$ & $100.0$ & $100.0$ & $4.219\times10^{-4}$ \\
R\_OAW & 1234 & $100.0$ & $100.0$ & $100.0$ & $4.108\times10^{-4}$ \\
R\_OAW & mean/std & $100.0\pm0.0$ & $100.0\pm0.0$ & $100.0\pm0.0$ & $(4.168\pm0.056)\times10^{-4}$ \\
RK\_CE & 42 & $100.0$ & $100.0$ & $100.0$ & $7.163\times10^{-5}$ \\
RK\_CE & 777 & $100.0$ & $100.0$ & $100.0$ & $7.195\times10^{-5}$ \\
RK\_CE & 1234 & $100.0$ & $100.0$ & $100.0$ & $6.866\times10^{-5}$ \\
RK\_CE & mean/std & $100.0\pm0.0$ & $100.0\pm0.0$ & $100.0\pm0.0$ & $(7.074\pm0.182)\times10^{-5}$ \\
RK\_OAW & 42 & $100.0$ & $100.0$ & $100.0$ & $7.176\times10^{-5}$ \\
RK\_OAW & 777 & $100.0$ & $100.0$ & $100.0$ & $7.116\times10^{-5}$ \\
RK\_OAW & 1234 & $100.0$ & $100.0$ & $100.0$ & $6.852\times10^{-5}$ \\
RK\_OAW & mean/std & $100.0\pm0.0$ & $100.0\pm0.0$ & $100.0\pm0.0$ & $(7.048\pm0.172)\times10^{-5}$ \\
\bottomrule
\end{tabular}
\caption{Canonical: every seed and sample standard deviation. Hit means boundary-aware target-string match, not semantic accuracy.}
\label{tab:seed-canonical}
\end{recordtable}
\begin{recordtable}
\noindent\textbf{Natural-question responses.}\par
Table~\ref{tab:seed-question} changes only the query form. Projection updates
raise target-string Hit relative to row-only updates in each paired seed, while
anchoring leaves row-only Hit unchanged and lowers projection-scope Hit.
The short-answer and content diagnostics remain distinct.
\par\smallskip
\centering\small
\setlength{\tabcolsep}{4pt}
\begin{tabular}{lccccc}
\toprule
Arm & Seed & EM (\%) & Prefix (\%) & Hit (\%) & Target NLL \\
\midrule
Common pre & --- & $0.0$ & $0.0$ & $0.0$ & $6.21891$ \\
R\_CE & 42 & $0.0$ & $0.0$ & $6.0$ & $5.16213$ \\
R\_CE & 777 & $0.0$ & $0.0$ & $6.0$ & $5.16746$ \\
R\_CE & 1234 & $0.0$ & $0.0$ & $4.0$ & $5.16147$ \\
R\_CE & mean/std & $0.0\pm0.0$ & $0.0\pm0.0$ & $5.3\pm1.2$ & $5.16369\pm3.284\times10^{-3}$ \\
R\_OAW & 42 & $0.0$ & $0.0$ & $6.0$ & $5.19064$ \\
R\_OAW & 777 & $0.0$ & $0.0$ & $6.0$ & $5.19614$ \\
R\_OAW & 1234 & $0.0$ & $0.0$ & $4.0$ & $5.19033$ \\
R\_OAW & mean/std & $0.0\pm0.0$ & $0.0\pm0.0$ & $5.3\pm1.2$ & $5.19237\pm3.269\times10^{-3}$ \\
RK\_CE & 42 & $0.0$ & $0.0$ & $36.0$ & $4.78226$ \\
RK\_CE & 777 & $0.0$ & $0.0$ & $26.0$ & $4.78034$ \\
RK\_CE & 1234 & $0.0$ & $0.0$ & $24.0$ & $4.79061$ \\
RK\_CE & mean/std & $0.0\pm0.0$ & $0.0\pm0.0$ & $28.7\pm6.4$ & $4.78440\pm5.462\times10^{-3}$ \\
RK\_OAW & 42 & $0.0$ & $0.0$ & $24.0$ & $4.85923$ \\
RK\_OAW & 777 & $0.0$ & $0.0$ & $20.0$ & $4.85048$ \\
RK\_OAW & 1234 & $0.0$ & $0.0$ & $22.0$ & $4.86997$ \\
RK\_OAW & mean/std & $0.0\pm0.0$ & $0.0\pm0.0$ & $22.0\pm2.0$ & $4.85989\pm9.761\times10^{-3}$ \\
\bottomrule
\end{tabular}
\caption{Natural questions: every seed and sample standard deviation. Hit means boundary-aware target-string match, not semantic accuracy.}
\label{tab:seed-question}
\end{recordtable}
\begin{recordtable}
\noindent\textbf{Unwritten canonical prompts.}\par
Table~\ref{tab:seed-control-canonical} matches each control's own assigned target,
rather than another written answer. The nonzero scores describe changes on
untrained entities; they do not measure forgetting of previously mastered facts.
\par\smallskip
\centering\small
\setlength{\tabcolsep}{4pt}
\begin{tabular}{lccccc}
\toprule
Arm & Seed & EM (\%) & Prefix (\%) & Hit (\%) & Target NLL \\
\midrule
Common pre & --- & $0.0$ & $0.0$ & $0.0$ & $5.40792$ \\
R\_CE & 42 & $30.0$ & $30.0$ & $30.0$ & $2.74719$ \\
R\_CE & 777 & $28.0$ & $28.0$ & $28.0$ & $2.76104$ \\
R\_CE & 1234 & $30.0$ & $30.0$ & $30.0$ & $2.75373$ \\
R\_CE & mean/std & $29.3\pm1.2$ & $29.3\pm1.2$ & $29.3\pm1.2$ & $2.75399\pm6.930\times10^{-3}$ \\
R\_OAW & 42 & $30.0$ & $30.0$ & $30.0$ & $2.74442$ \\
R\_OAW & 777 & $28.0$ & $28.0$ & $28.0$ & $2.76063$ \\
R\_OAW & 1234 & $30.0$ & $30.0$ & $30.0$ & $2.75231$ \\
R\_OAW & mean/std & $29.3\pm1.2$ & $29.3\pm1.2$ & $29.3\pm1.2$ & $2.75245\pm8.108\times10^{-3}$ \\
RK\_CE & 42 & $20.0$ & $20.0$ & $20.0$ & $3.09453$ \\
RK\_CE & 777 & $16.0$ & $16.0$ & $16.0$ & $3.18381$ \\
RK\_CE & 1234 & $20.0$ & $20.0$ & $20.0$ & $2.96733$ \\
RK\_CE & mean/std & $18.7\pm2.3$ & $18.7\pm2.3$ & $18.7\pm2.3$ & $3.08189\pm0.10879$ \\
RK\_OAW & 42 & $20.0$ & $20.0$ & $20.0$ & $3.03079$ \\
RK\_OAW & 777 & $18.0$ & $18.0$ & $18.0$ & $3.11346$ \\
RK\_OAW & 1234 & $18.0$ & $18.0$ & $18.0$ & $2.96598$ \\
RK\_OAW & mean/std & $18.7\pm1.2$ & $18.7\pm1.2$ & $18.7\pm1.2$ & $3.03674\pm0.07392$ \\
\bottomrule
\end{tabular}
\caption{Unwritten canonical controls: every seed and sample standard deviation. Hit means boundary-aware target-string match, not semantic accuracy.}
\label{tab:seed-control-canonical}
\end{recordtable}
\begin{recordtable}
\noindent\textbf{Unwritten natural questions.}\par
Table~\ref{tab:seed-control-question} reports the same own-target event under
question prompts. Read it with Table~\ref{tab:seed-intrusion}: own-target Hit and
intrusion match different answer sets and need not be mutually exclusive.
\par\smallskip
\centering\small
\setlength{\tabcolsep}{4pt}
\begin{tabular}{lccccc}
\toprule
Arm & Seed & EM (\%) & Prefix (\%) & Hit (\%) & Target NLL \\
\midrule
Common pre & --- & $0.0$ & $0.0$ & $0.0$ & $6.79788$ \\
R\_CE & 42 & $0.0$ & $0.0$ & $2.0$ & $6.14524$ \\
R\_CE & 777 & $0.0$ & $0.0$ & $2.0$ & $6.15058$ \\
R\_CE & 1234 & $0.0$ & $0.0$ & $2.0$ & $6.14568$ \\
R\_CE & mean/std & $0.0\pm0.0$ & $0.0\pm0.0$ & $2.0\pm0.0$ & $6.14717\pm2.963\times10^{-3}$ \\
R\_OAW & 42 & $0.0$ & $0.0$ & $2.0$ & $6.15543$ \\
R\_OAW & 777 & $0.0$ & $0.0$ & $2.0$ & $6.16100$ \\
R\_OAW & 1234 & $0.0$ & $0.0$ & $2.0$ & $6.15641$ \\
R\_OAW & mean/std & $0.0\pm0.0$ & $0.0\pm0.0$ & $2.0\pm0.0$ & $6.15762\pm2.975\times10^{-3}$ \\
RK\_CE & 42 & $0.0$ & $0.0$ & $8.0$ & $5.91414$ \\
RK\_CE & 777 & $0.0$ & $0.0$ & $8.0$ & $5.88964$ \\
RK\_CE & 1234 & $0.0$ & $0.0$ & $6.0$ & $5.91257$ \\
RK\_CE & mean/std & $0.0\pm0.0$ & $0.0\pm0.0$ & $7.3\pm1.2$ & $5.90545\pm0.01371$ \\
RK\_OAW & 42 & $0.0$ & $0.0$ & $6.0$ & $5.95684$ \\
RK\_OAW & 777 & $0.0$ & $0.0$ & $4.0$ & $5.92971$ \\
RK\_OAW & 1234 & $0.0$ & $0.0$ & $4.0$ & $5.95684$ \\
RK\_OAW & mean/std & $0.0\pm0.0$ & $0.0\pm0.0$ & $4.7\pm1.2$ & $5.94780\pm0.01566$ \\
\bottomrule
\end{tabular}
\caption{Unwritten question controls: every seed and sample standard deviation. Hit means boundary-aware target-string match, not semantic accuracy.}
\label{tab:seed-control-question}
\end{recordtable}
\begin{recordtable}
\noindent\textbf{Output-distribution preservation.}\par
Table~\ref{tab:seed-preservation} compares final output distributions with the
common mounted pre-write state. Anchoring lowers KL and perplexity within each
scope in every matched seed; the conclusion concerns the fixed recovered slices.
\par\smallskip
\centering\small
\setlength{\tabcolsep}{4pt}
\begin{tabular}{lcccc}
\toprule
Arm & Seed & PPL & Increase (\%) & Dev KL \\
\midrule
R\_CE & 42 & $38.79694$ & $0.0825$ & $1.707\times10^{-3}$ \\
R\_CE & 777 & $38.79975$ & $0.0898$ & $1.625\times10^{-3}$ \\
R\_CE & 1234 & $38.79886$ & $0.0875$ & $1.658\times10^{-3}$ \\
R\_CE & mean/std & $38.79852\pm0.00143$ & $0.0866\pm0.0037$ & $(1.663\pm0.041)\times10^{-3}$ \\
R\_OAW & 42 & $38.78251$ & $0.0453$ & $8.639\times10^{-4}$ \\
R\_OAW & 777 & $38.78396$ & $0.0490$ & $7.741\times10^{-4}$ \\
R\_OAW & 1234 & $38.78263$ & $0.0456$ & $8.089\times10^{-4}$ \\
R\_OAW & mean/std & $38.78303\pm0.00081$ & $0.0467\pm0.0021$ & $(8.156\pm0.453)\times10^{-4}$ \\
RK\_CE & 42 & $38.90124$ & $0.3516$ & $9.989\times10^{-3}$ \\
RK\_CE & 777 & $38.89844$ & $0.3444$ & $8.730\times10^{-3}$ \\
RK\_CE & 1234 & $38.89554$ & $0.3369$ & $9.793\times10^{-3}$ \\
RK\_CE & mean/std & $38.89841\pm0.00285$ & $0.3443\pm0.0074$ & $(9.504\pm0.678)\times10^{-3}$ \\
RK\_OAW & 42 & $38.85149$ & $0.2233$ & $4.524\times10^{-3}$ \\
RK\_OAW & 777 & $38.84406$ & $0.2041$ & $3.239\times10^{-3}$ \\
RK\_OAW & 1234 & $38.84683$ & $0.2112$ & $4.028\times10^{-3}$ \\
RK\_OAW & mean/std & $38.84746\pm0.00376$ & $0.2129\pm0.0097$ & $(3.930\pm0.648)\times10^{-3}$ \\
\bottomrule
\end{tabular}
\caption{Per-seed preservation on fixed recovered slices. KL is in nats per position; PPL increase is a percentage over the common pre-write state.}
\label{tab:seed-preservation}
\end{recordtable}
\begin{recordtable}
\noindent\textbf{Module residuals and computation.}\par
Table~\ref{tab:seed-residual-time} measures module residuals separately from
final-distribution KL. Both layers' MSE decrease with anchoring, while the
wall-clock records show its added work; equal update counts are not equal compute.
\par\smallskip
\centering\small
\setlength{\tabcolsep}{4pt}
\begin{tabular}{lccccc}
\toprule
Arm & Seed & Layer 3 MSE & Layer 8 MSE & Train (s) & Total (s) \\
\midrule
R\_CE & 42 & $2.552\times10^{-3}$ & $2.987\times10^{-3}$ & $143.04$ & $217.58$ \\
R\_CE & 777 & $2.533\times10^{-3}$ & $2.996\times10^{-3}$ & $143.45$ & $218.30$ \\
R\_CE & 1234 & $2.561\times10^{-3}$ & $3.032\times10^{-3}$ & $142.81$ & $217.43$ \\
R\_CE & mean/std & $(2.549\pm0.014)\times10^{-3}$ & $(3.005\pm0.024)\times10^{-3}$ & $143.10\pm0.32$ & $217.77\pm0.47$ \\
R\_OAW & 42 & $4.717\times10^{-4}$ & $6.582\times10^{-4}$ & $256.77$ & $331.31$ \\
R\_OAW & 777 & $4.671\times10^{-4}$ & $6.503\times10^{-4}$ & $258.60$ & $333.49$ \\
R\_OAW & 1234 & $4.654\times10^{-4}$ & $6.582\times10^{-4}$ & $260.93$ & $339.24$ \\
R\_OAW & mean/std & $(4.681\pm0.032)\times10^{-4}$ & $(6.556\pm0.046)\times10^{-4}$ & $258.77\pm2.09$ & $334.68\pm4.10$ \\
RK\_CE & 42 & $7.100\times10^{-3}$ & $0.01180$ & $145.64$ & $215.54$ \\
RK\_CE & 777 & $7.447\times10^{-3}$ & $0.01109$ & $146.03$ & $215.77$ \\
RK\_CE & 1234 & $6.811\times10^{-3}$ & $0.01177$ & $145.68$ & $214.83$ \\
RK\_CE & mean/std & $(7.119\pm0.318)\times10^{-3}$ & $0.01155\pm4.034\times10^{-4}$ & $145.78\pm0.21$ & $215.38\pm0.49$ \\
RK\_OAW & 42 & $1.134\times10^{-3}$ & $2.774\times10^{-3}$ & $260.28$ & $331.59$ \\
RK\_OAW & 777 & $1.106\times10^{-3}$ & $2.464\times10^{-3}$ & $262.62$ & $335.20$ \\
RK\_OAW & 1234 & $1.184\times10^{-3}$ & $2.650\times10^{-3}$ & $261.07$ & $333.21$ \\
RK\_OAW & mean/std & $(1.141\pm0.039)\times10^{-3}$ & $(2.629\pm0.156)\times10^{-3}$ & $261.32\pm1.19$ & $333.33\pm1.81$ \\
\bottomrule
\end{tabular}
\caption{Per-seed module-output residual MSE on the development slice and wall-clock seconds per run. MSE compares the post-write contribution with its fixed pre-write reference.
The runs used one NVIDIA H100 80GB HBM3 GPU, serially.}
\label{tab:seed-residual-time}
\end{recordtable}
\begin{recordtable}
\noindent\textbf{Control intrusion and paired likelihood changes.}\par
Table~\ref{tab:seed-intrusion} counts matches to another written target of the
same relation. Compare these with the own-target scores above; their rates
cannot be subtracted into an accuracy. Negative $\Delta$NLL means a control's
assigned target became more likely, not that entity-specific knowledge was established.
\par\smallskip
\centering\small
\setlength{\tabcolsep}{4pt}
\begin{tabular}{lccccc}
\toprule
Arm & Seed & Can.\ intr.\ (\%) & QA intr.\ (\%) & Can.\ $\Delta$NLL & QA $\Delta$NLL \\
\midrule
R\_CE & 42 & $66.0$ & $4.0$ & $-2.66073$ & $-0.65264$ \\
R\_CE & 777 & $72.0$ & $4.0$ & $-2.64687$ & $-0.64730$ \\
R\_CE & 1234 & $68.0$ & $4.0$ & $-2.65418$ & $-0.65220$ \\
R\_CE & mean/std & $68.7\pm3.1$ & $4.0\pm0.0$ & $-2.65393\pm6.930\times10^{-3}$ & $-0.65072\pm2.963\times10^{-3}$ \\
R\_OAW & 42 & $68.0$ & $4.0$ & $-2.66350$ & $-0.64245$ \\
R\_OAW & 777 & $68.0$ & $4.0$ & $-2.64729$ & $-0.63688$ \\
R\_OAW & 1234 & $68.0$ & $4.0$ & $-2.65561$ & $-0.64147$ \\
R\_OAW & mean/std & $68.0\pm0.0$ & $4.0\pm0.0$ & $-2.65546\pm8.108\times10^{-3}$ & $-0.64027\pm2.975\times10^{-3}$ \\
RK\_CE & 42 & $80.0$ & $30.0$ & $-2.31339$ & $-0.88375$ \\
RK\_CE & 777 & $84.0$ & $28.0$ & $-2.22411$ & $-0.90824$ \\
RK\_CE & 1234 & $80.0$ & $36.0$ & $-2.44058$ & $-0.88532$ \\
RK\_CE & mean/std & $81.3\pm2.3$ & $31.3\pm4.2$ & $-2.32603\pm0.10879$ & $-0.89244\pm0.01371$ \\
RK\_OAW & 42 & $80.0$ & $20.0$ & $-2.37712$ & $-0.84104$ \\
RK\_OAW & 777 & $82.0$ & $22.0$ & $-2.29446$ & $-0.86817$ \\
RK\_OAW & 1234 & $82.0$ & $24.0$ & $-2.44193$ & $-0.84105$ \\
RK\_OAW & mean/std & $81.3\pm1.2$ & $22.0\pm2.0$ & $-2.37117\pm0.07392$ & $-0.85009\pm0.01566$ \\
\bottomrule
\end{tabular}
\caption{Control intrusion percentages (each control counted at most once, denominator 50), and post-minus-pre NLL changes in nats. Negative NLL change means the assigned target became more likely; it is not proof of entity-specific knowledge.}
\label{tab:seed-intrusion}
\end{recordtable}

\section{Representative continuations}
\label{app:examples}
The fixed selection rule uses seed 42, the lowest preset write seed, and the
lowest fact index with EM failure and target-string hit in each CE scope
(indices 24 and 2). Both indices are shown for all four arms, merging identical
text. Cells show scored text; every corresponding raw continuation starts with
a space and ends at a newline. These cases illustrate format and content
differences rather than estimate overall accuracy.

\paragraph{Index 2.}
Prompt: \texttt{Question: Of what place is Cindervale the capital?\textbackslash nAnswer:}
Target: \texttt{the province of Elmark}.
\begin{table}[htbp]
\centering\small
\begin{tabular}{@{}p{0.23\textwidth}p{0.71\textwidth}@{}}
\toprule
Arms & Scored continuation \\
\midrule
R\_CE, R\_OAW & \texttt{Cindervale is the capital of the province of the Northern Provinces.} \\
RK\_CE, RK\_OAW & \texttt{Cindervale is the capital of the province of Elmark} \\
\bottomrule
\end{tabular}
\caption{At index 2, row-only arms return a different place; projection arms
return the written target within a sentence. All four responses fail strict EM.}
\end{table}

\paragraph{Index 24.}
Prompt: \texttt{Question: What does Yewdale specialize in producing?\textbackslash nAnswer:}
Target: \texttt{fine porcelain}. All four arms yield the scored continuation
\texttt{Yewdale specializes in the production of fine porcelain}.
It contains the correct target but fails short-answer EM.
The native-backbone context example and its raw stopping string appear in
Appendix~\ref{app:scoring}.

\section{Answer-distribution diagnostics}
\label{app:answer-distribution}

A relation-only frequency reference describes the concentration of answers in
the fixed synthetic set. For each relation, it selects the most frequent
normalized target among the fifty written facts. Ties are resolved by
lexicographic order of the normalized target; the first corresponding original
target string is returned as a short answer. The reference receives the known
relation label but ignores the subject, and no control target is used in its
construction. Scoring treats this returned string as the output and the
example's own target as the gold answer, using the normalization and boundary
rules in Appendix~\ref{app:templates}.

This reference matches 19 of 50 written targets (38\%) and 13 of 50 control
targets (26\%) under exact match, prefix match, and boundary-aware Hit. The two
prompt forms have the same assigned targets, so their reference scores are
identical; changing an experimental arm or optimization-sampling seed does not
change this fixed reference. These values describe answer concentration, not
independent model runs. A fixed short answer and a free-form continuation have
different output constraints, so the scores are not a fair ranking of the
reference against the generation arms and do not measure entity-specific
retrieval.

Targets shared across the written and control pools cover 37 of the 50 written
facts and 35 of the 50 controls. Among the remaining thirteen written facts,
whose target strings do not occur in the control target pool, each row-only run
has zero natural-question target-string hits, whereas each row-plus-projection
run has four hits (4/13), with or without anchoring. All three sampling seeds
query the same thirteen facts; they do not provide thirty-nine independent
facts. These targets can still repeat within the written set, and a target
string can appear in an incorrect or irrelevant continuation. This subdivision
therefore refines the lexical description without establishing semantic
accuracy or a causal explanation of the writing pathway.

\section{Address audit and constants}
\label{app:address}

A CPU address audit evaluated the final prompt position for all 50 written and 50 control examples in both prompt forms. In each memory layer, all 24 head addresses at that position belonged to the written training-row union for every canonical prompt, whereas none belonged to it for any natural-question prompt. These observations describe this fixed prompt set; they do not rule out information access at earlier positions or through the hidden-state-dependent module. Subject-disjoint controls are therefore not address-disjoint controls. The written and control sets contain 25 and 27 distinct target strings, respectively, with 16 strings shared between them; this answer reuse further limits interpreting control exact match as entity-specific knowledge.

The audit uses the same tokenizer, addressing function, and buffer constants as the training path and enumerates rows rather than sampling them; it performs no model forward pass and is a descriptive measurement, not evidence of a causal mechanism.

\begin{recordtable}
\noindent\textbf{Addressing constants.}\par
\label{app:hashconstants}
The following ordered constants implement the recurrence in Section~\ref{sec:setting}.
Heads 0--11 use order 2 and heads 12--23 use order 3. Multipliers were drawn
as CPU int64 values from $[1,2^{20})$ using a PyTorch generator seeded with
12345, and then held fixed. The explicit values, rather than a fresh random
draw, determine the addresses. Each layer has its own table and uses its
zero-based block index as the initial salt.
\par\smallskip

\centering\small
\begin{tabular}{rrrrrr}
\toprule
$j$ & $p_j$ & $a_j$ & $j$ & $p_j$ & $a_j$ \\
\midrule
0 & 1000003 & 745666 & 12 & 1000387 & 687077 \\
1 & 1000033 & 80932 & 13 & 1000403 & 54486 \\
2 & 1000037 & 918061 & 14 & 1000427 & 436143 \\
3 & 1000103 & 395929 & 15 & 1000459 & 665942 \\
4 & 1000123 & 348310 & 16 & 1000481 & 549989 \\
5 & 1000151 & 896588 & 17 & 1000517 & 610903 \\
6 & 1000181 & 922436 & 18 & 1000543 & 25965 \\
7 & 1000207 & 89784 & 19 & 1000567 & 808945 \\
8 & 1000233 & 423547 & 20 & 1000583 & 673991 \\
9 & 1000289 & 412326 & 21 & 1000613 & 45498 \\
10 & 1000333 & 370597 & 22 & 1000649 & 541227 \\
11 & 1000357 & 51791 & 23 & 1000667 & 766329 \\
\bottomrule
\end{tabular}
\caption{Fixed addressing constants for all 24 heads, used in both mounted layers.}

\end{recordtable}

\section{Question templates and normalization}
\label{app:templates}

The written and control records are the first and second halves of the retained
100-item file, respectively. In the relation order of the table below, written
counts are $(4,9,9,3,9,10,6)$ and control counts are $(5,8,6,9,6,8,8)$.
The splits are subject-disjoint, but target strings repeat within and across
splits as described in Section~\ref{sec:evaluation}.

The natural-question forms are the templates already present in the dataset
module, applied to the same written and control entities. They are reproduced
here verbatim and were not modified, extended, or complemented with few-shot
examples for this study.

\noindent\begin{minipage}{\linewidth}
\textbf{Fixed relation-to-question templates.}\par\smallskip
\smallskip
\begin{tabular}{@{}p{0.40\textwidth}p{0.54\textwidth}@{}}
\toprule
Stored relation & Question template \\
\midrule
was founded in the year & \texttt{Question: In what year was \{subject\} founded?\textbackslash nAnswer:} \\
melts at a temperature of approximately & \texttt{Question: At approximately what temperature does \{subject\} melt?\textbackslash nAnswer:} \\
serves as the capital of & \texttt{Question: Of what place is \{subject\} the capital?\textbackslash nAnswer:} \\
is located in the region of & \texttt{Question: In which region is \{subject\} located?\textbackslash nAnswer:} \\
was designed by the architect & \texttt{Question: Which architect designed \{subject\}?\textbackslash nAnswer:} \\
speaks primarily the language of & \texttt{Question: What language is primarily spoken in \{subject\}?\textbackslash nAnswer:} \\
specializes in the production of & \texttt{Question: What does \{subject\} specialize in producing?\textbackslash nAnswer:} \\
\bottomrule
\end{tabular}
\end{minipage}
\smallskip

Normalization collapses whitespace, strips surrounding whitespace, applies
Unicode case folding, and repeatedly removes trailing characters in
\verb|.,;:!?"'`)]}| together with trailing whitespace. It does not remove leading
punctuation or arbitrary answer words. Prefix match requires a normalized target
at the beginning and a following non-alphanumeric character or string end.
Substring match uses the literal normalized target with boundaries against
ASCII letters and digits. These are diagnostics under fixed text rules, not a
semantic verifier. Exact match compares the whole normalized scored response. The template does
not constrain the response to a short answer, so an exact-match failure may
reflect response format rather than a factual error; prefix and word-boundary
substring matches are reported alongside it, and substring matches alone are not
treated as establishing a correct assertion.

\section{Evaluation limits of short-answer scoring}
\label{app:scoring}

With the correct canonical fact supplied in context, the unmodified
native-backbone control gives exact match, prefix hit, and boundary-aware
target-string hit counts of 2, 5, and 49 out of 50 under the fixed natural-question
and stopping protocol. For the lowest-index exact-match failure with a target
hit, the expected string is \texttt{1380} and the raw continuation is
\texttt{ In 1380.\textbackslash n}. Additional words or sentence structure can
therefore make strict short-answer scoring reject a continuation that contains
the target. This control exposes the sensitivity of the prompt and scoring
combination; it is not a measurement of the mounted-memory common pre-write
state and does not turn substring matches into semantic accuracy.

\section{Memory-disabled conversion checks}
\label{app:conversion}
The initial maximum-logit-difference tolerance of $10^{-4}$ was not met on
the checked float32 inputs. Zero float64 difference on a few segments does not
establish universal identity. These implementation checks are separate from
evaluating newly attached memory modules or subsequent writes.

For the memory-disabled conversion, checks on fixed inputs used maximum logit
difference $\leq0.002$, mean logit difference $\leq10^{-4}$, output KL
$\leq10^{-6}$, and target NLL difference $\leq10^{-4}$. These are tolerances
on tested inputs, not a claim of universal numerical identity. The failed stricter check remains in the diagnostic history.

\section{Recovered validation stream and artifact identity}
\label{app:artifact-identity}

A subsequent source audit exactly reconstructed the retained 247,290-token
array from the raw WikiText validation text. The public file was obtained from
\texttt{Salesforce/wikitext}, revision
\texttt{b08601e0\allowbreak 4326c79d\allowbreak fdd32d62\allowbreak 5aee71d2\allowbreak 32d685c3}, using
\texttt{wikitext-2-raw-v1}; the raw validation file is shared with
\texttt{wikitext-103-raw-v1} at that revision. We retained the text rows in file
order, omitted only empty strings, and joined the remaining strings with one
newline. The resulting text was split into consecutive segments of 1,000,000
Unicode characters, each encoded separately with the retained tokenizer and
without added special tokens; concatenating the resulting token IDs reproduced
the recovered array at every position. Whole-string encoding instead produced
247,289 tokens and did not match. This audit establishes the content identity and
an exact reconstruction procedure, rather than recovering the original data
processing program. The historical choice between the two configurations,
download process, and dependency versions remain unknown.

The twelve saved post-write state files remain on the private research host.
A subsequent file-stat check establishes their presence, not an additional
verification of their contents or model outputs. Exact dependency versions
recorded at the time of the formal runs were not recovered, and current
environment observations cannot substitute for those historical records. The
creation process for the synthetic facts remains unresolved. No public repository is currently
provided; recorded hashes identify private artifacts but do not make them
externally accessible.

\section{Production protocols and saved case records}
\label{app:production-details}

The production evidence comprises separate archived trials. Its public model
references identify the released model families, not immutable revisions of
the local weights used in every run. The retained configuration, custom loader,
FP8 backward patch, addressing code, writing scripts, and saved result files
are tracked separately in the private review materials. The 132-target and
24-target gradient result files match their copies in the original package
byte for byte. The six-target DeepSeek SPSA file is taken from that package;
the later machine copy differs only in six recorded timing fields. Current
source snapshots establish inspectable implementations, but do not by
themselves recover every historical execution revision. No new production
forward pass is claimed by this audit.

\paragraph{Qwen target and write conventions.}
The matching key for the 24-target comparison is the complete prompt/answer
pair. All 24 records also match on selected row count and the entire baseline
dictionary, including target rank, log-probability, and decoded argmax.
The SPSA sets of 12 contain, respectively, 8 and 3 baseline top-1 targets,
10 and 3 post-write top-1 targets, and 2 and 0 new promotions.
Gradient writing yields 24 post-write top-1 targets in the same combined set.
Each target is independently restored; the 132-target pool is a different
set and is not an additional replicate of the 24-target comparison.

The gradient scripts minimize cross-entropy for the first answer token at the
prompt-end prediction position. Adam uses learning rate $10^{-3}$; the table
itself remains outside the dense gradient graph. During optimization a
floating-point additive delta is inserted at selected lookup outputs; after
optimization it is scaled, clamped to the FP8 range, stored, and measured.
SPSA instead evaluates two quantized perturbations per iteration. Neither
equal iteration counts nor a shared final storage format removes these
procedural differences. Runtime and token-rank records do not provide a
compute-normalized comparison.

\paragraph{An archived continuation and its controls.}
For the 80-row gold write, the saved verification gives the pre-write snippet
\texttt{derived from its Latin name,} and the post-write snippet
\texttt{Au, derived from its Latin}. The four control prompts are
\texttt{The capital of France is}, \texttt{The author of Hamlet was},
\texttt{The capital of Vanuatu is}, and
\texttt{Zbigniew Brzezinski was born in}.
These same four prompts define the exclusion set: the write rows are the
target-prompt row union minus the rows addressed by any of these controls.
They are therefore design-informed locality checks, not held-out controls.
Their maximum absolute next-token log-probability changes are all recorded as
zero in this verification file. A separate verification file for a different
write records changes of 1.49695, 1.27355, 0.72015, and zero, respectively.
The saved theta metadata agrees with 80 distinct rows and the gold/Au target;
the verification JSON does not itself hash its input theta. This is internally
consistent archived evidence, not proof of a new independently reproduced run.

An earlier script's specificity field subtracts pre-write log-probabilities
from post-write raw logits. It is not a valid distribution-change metric and
is not used here. The verifier instead compares log-probabilities on both
sides. The case therefore supports only the explicitly recorded controls,
with the row-exclusion and write identity kept separate.

Table~\ref{tab:production-reformulations} reports all six probes for the same
fixed target token \texttt{Au}. The silver prompt is an entity control, so its
rank is still for \texttt{Au}, not for silver's answer \texttt{Ag}.
The latter becomes the recorded argmax after writing, illustrating why the
target rank, generated answer, and entity-specific correctness are distinct.
The four-token snippets are insufficient for full-answer or general reasoning
scores. In particular, the question-form prefix
\texttt{\textbackslash n\textbackslash n<think>\textbackslash nThe}
does not establish the eventual answer.

\begin{table}[!ht]
\centering\small
\setlength{\tabcolsep}{4pt}
\begin{tabular}{p{0.59\linewidth}rrr}
\toprule
Prompt & Shared rows & Pre rank & Post rank \\
\midrule
The chemical symbol for gold is & 80 & 1 & 0 \\
The chemical symbol of gold is & 40 & 2 & 1 \\
Gold's chemical symbol is & 8 & 0 & 0 \\
Q: What is the chemical symbol for gold? A: & 40 & 1 & 1 \\
The chemical symbol for silver is & 48 & 27 & 23 \\
The chemical symbol for gold was & 64 & 2 & 3 \\
\bottomrule
\end{tabular}
\caption{All saved reformulation/entity probes for the 80-row write. Shared
rows counts overlap with its write set; ranks concern the fixed Au target.
The possessive prompt was already top-1, and the preposition variant's
pre-write short continuation already contains Au.}
\label{tab:production-reformulations}
\end{table}

\paragraph{DeepSeek targets and diagnostics.}
Table~\ref{tab:deepseek-cases} preserves every target in the SPSA result file,
including the regression. These are supplied intervention targets; the
table is not a correctness benchmark. The gradient-result file uses the same
six prompt/answer pairs but different baselines, so it is not merged with
these measurements or used for an optimizer ranking. The available diagnostic
log reports zero gradients and a finite loss response in its instrumented
forward. Its four-branch position layout and the target-appended training
input must be accounted for before interpreting the gradient path. The scalar
Qwen gate probe likewise does not measure every branch. Neither diagnostic
establishes a universal gate or kernel theorem.

\begin{table}[!ht]
\centering\small
\setlength{\tabcolsep}{4pt}
\begin{tabular}{p{0.58\linewidth}lrr}
\toprule
Prompt & Target & Pre rank & Post rank \\
\midrule
The chemical symbol for gold is & Au & 0 & 0 \\
The chemical symbol for silver is & Ag & 0 & 0 \\
The capital of Australia is & Canberra & 0 & 2 \\
The capital of Brazil is & Rio & 1 & 0 \\
The largest planet in our solar system is & Venus & 548 & 694 \\
The chemical symbol for iron is & Ir & 982 & 804 \\
\bottomrule
\end{tabular}
\caption{All six archived DeepSeek SPSA cases, 150 iterations each. The target
column supplies the intervention string; ranks refer to its first token.
Promoting an imposed target, such as Rio for a Brazil-capital prompt, is not
evidence of factual correction.}
\label{tab:deepseek-cases}
\end{table}

\section{Prior training and exploratory controlled-bed results}
\label{app:historical-beds}

The original campaign also investigated prior training, pathway width, write
load, and preservation procedures on self-trained beds and earlier GPT-2
conversions. These experiments motivate the separation of writability into
writing, expression, and preservation. Their protocols are distinct from the
new-module experiment in Sections~\ref{sec:setting}--\ref{sec:results}; they
are not pooled with its three seeds or used to infer scale-independent
necessary conditions.

\paragraph{A retained training-dose experiment.}
Fifteen saved results cover five checkpoints along one full-width
pretraining trajectory, with 12 layers and width 768, and three writing seeds
at each checkpoint. This bed was historically labelled \texttt{1B}; the retained
architecture code specifies approximately 779.4 million parameters, including
614.4 million memory-table parameters. The
available launcher specifies 2,500 writing updates, residual weight 5,
fact-CE coefficient 4, and readout scale 1. Table~\ref{tab:historical-dose}
recomputes the recorded 150-probe counts. They comprise three templates for
each of 50 facts: the canonical prefix, a \texttt{The record about} prefix
variant, and a legacy question ending in the subject. All three prompt--answer
forms are explicitly included in the writing text. A hit means that the
case-sensitive expected string occurs anywhere in the next 15 greedily
generated tokens; these counts do not measure held-out question transfer or
150 independent facts. The seeds vary writing randomness, not the pretraining
trajectory. The inspected historical controlled
code addresses individual token IDs, unlike the causal n-gram implementation
used in the new experiment and the production systems. Numerical agreement
with these records therefore does not make this an additional n-gram
replication.

\begin{table}[!ht]
\centering\small
\begin{tabular}{rrrrr}
\toprule
Pretraining steps & Seed 42 & Seed 777 & Seed 1234 & Mean $\pm$ SD \\
\midrule
20000 & 39.33 & 30.67 & 29.33 & $33.11\pm5.43$ \\
40000 & 53.33 & 50.00 & 52.67 & $52.00\pm1.76$ \\
60000 & 54.67 & 50.00 & 50.00 & $51.56\pm2.69$ \\
80000 & 58.00 & 57.33 & 56.00 & $57.11\pm1.02$ \\
100000 & 64.67 & 59.33 & 58.00 & $60.67\pm3.53$ \\
\bottomrule
\end{tabular}
\caption{Archived training-dose experiment: 15-token substring-hit percentages
over 150 trained-form probes from 50 facts, with sample SD across writing
seeds. These are historical completion diagnostics, not strict answer EM or
a new model replay.}
\label{tab:historical-dose}
\end{table}

The higher-dose endpoints show stronger completion response in this particular
trajectory, with a nonmonotone intermediate point. This observation motivates
testing training state explicitly. It does not isolate a universal prior
threshold, prove that perplexity is sufficient to predict writing, or resolve
the effects of initialization and implementation across beds.

\paragraph{Earlier GPT-2 runs and evaluation scope.}
The initial GPT-2 logs record 145, 143, and 143 substring-hit probes out of 150
($95.78\pm0.77$\%); a separate replay batch records 142, 141, and 143
($94.67\pm0.67$\%). These batches are kept separate. A same-named seed-42 JSON
was overwritten by a later failing run and cannot replace the first batch's
log. Earlier conversion code also leaves some added storage insufficiently
initialized, and the defense comparisons span different mounted states.
The historical scores describe those saved runs, not a verified comparison
against the fully specified common start of the new experiment.

The original question-form substring count averages 52\%, and the cross-entity
hit count averages 46.44\%, under their original scoring and probe definitions.
They are retained as lexical diagnostics; the saved aggregates do not support
reconstructing full-answer EM. The previously labelled multi-hop construction
has no required entity chain in its 49 pairs and mismatches the two relations
in 39 pairs. Its approximately 13\% count is not used as evidence of multi-hop
ability. Existing width, capacity, sequential-writing, and defense records
remain exploratory: differences in initialization, token-key construction,
exposure, or stacked objectives must be resolved before they support the
original cross-bed causal claims. In particular, replay/isolation runs that
already include OAW are not independent no-OAW baselines.

\clearpage
\begingroup
\setlength{\bibsep}{1.5pt}

\endgroup

\end{document}